\documentclass[10pt,twocolumn,letterpaper]{article}

\usepackage{cvpr}              

\usepackage{graphicx}
\usepackage{amsmath}
\usepackage{amssymb}
\usepackage{booktabs}
\usepackage{multirow}
\usepackage[accsupp]{axessibility}  

\usepackage[pagebackref,breaklinks,colorlinks]{hyperref}

\usepackage[capitalize]{cleveref}
\crefname{section}{Sec.}{Secs.}
\Crefname{section}{Section}{Sections}
\Crefname{table}{Table}{Tables}
\crefname{table}{Tab.}{Tabs.}

\def\cvprPaperID{*****} 
\def\confName{CVPR}
\def\confYear{2022}

\begin{document}

\title{CoDo: Contrastive Learning with Downstream Background Invariance  for  Detection }

\author{Bing Zhao \quad Jun Li \quad Hong Zhu \\ 
Department of AI and HPC Inspur Electronic Information Industry Co., Ltd\\
Beijing, China\\
{\tt\small \{zhaobing01,lijun09,zhuhongbj\}@inspur.com}
}
\maketitle

\begin{abstract}
The prior self-supervised learning researches mainly select image-level instance discrimination as pretext task. It achieves a fantastic classification performance that is comparable to supervised learning methods. However, with degraded transfer performance on downstream tasks such as object detection. To bridge the performance gap, we propose a novel object-level self-supervised learning method, called \textbf{Co}ntrastive learning with \textbf{Do}wnstream background invariance (CoDo). The pretext task is converted to focus on instance location modeling for 
various backgrounds, especially for downstream datasets. The ability of background invariance is considered vital for object detection. Firstly, a data augmentation strategy is proposed to paste the instances onto background images, and then jitter the bounding box to involve background information. Secondly, we implement architecture alignment between our pretraining network and the mainstream detection pipelines. Thirdly, hierarchical  and multi views contrastive learning is designed to improve performance of visual representation learning. Experiments on MSCOCO demonstrate that the proposed CoDo with common backbones, ResNet50-FPN, yields strong transfer learning results for object detection.


\end{abstract}

\section{Introduction}
\label{sec:intro}

\begin{figure}
  \includegraphics[width=0.5\textwidth]{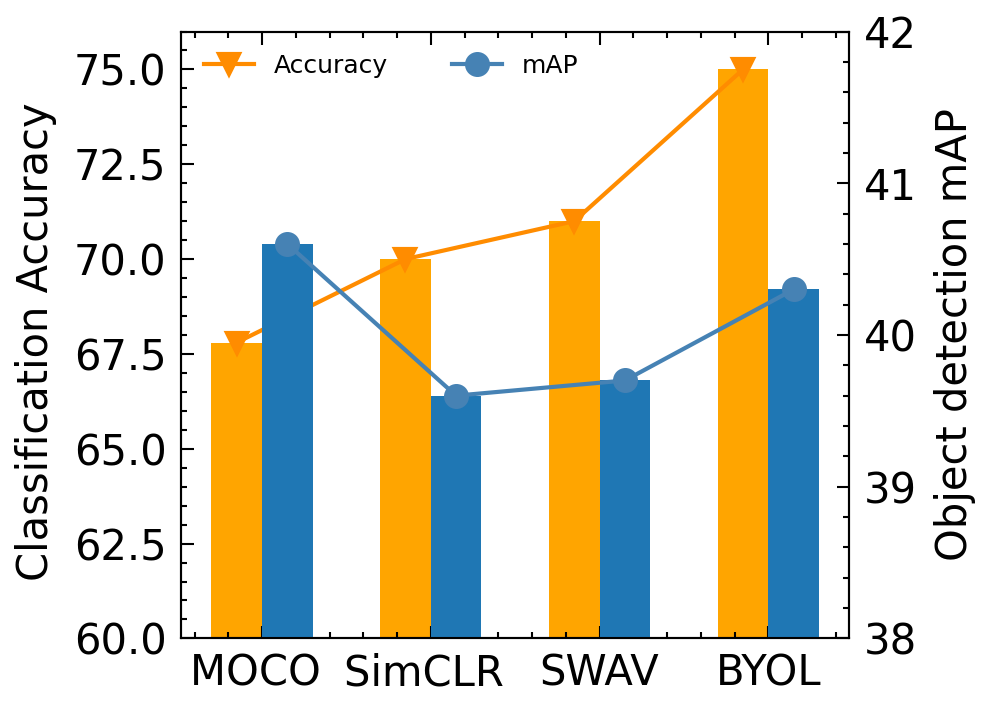}
  \caption{The performance inconsistency of related SSL methods in image classification and object detection. Specifically,   the classification performance of the relevant SSL methods on ImageNet is increasing, while there is no corresponding trend in object detection when finetuning the pre-trained weights on MSCOCO. } 
  
  \label{fig:1}
\end{figure}

The paradigm of supervised pretraining and finetuning has been dominant in computer vision for a period of time. Typically, the pretraining is optimized on large-scale labeled datasets, and then regarded as initialized weights to finetune for various downstream tasks~\cite{he2016deep}\cite{krizhevsky2012imagenet}\cite{zhuang2020comprehensive}. Instead, self-supervised learning (SSL) aims to learn the generic pretraining representations, independent of manual labels~\cite{liu2021self}. 
Recently, SSL has achieved a performance comparable to supervised pretraining in image classification~\cite{he2020momentum}\cite{chen2020simple}\cite{caron2020unsupervised}\cite{grill2020bootstrap}. However, it suffers a performance degradation when applied to downstream tasks, such as object detection. It indicates that the existing SSL approaches mainly focus on image classification, without considering the location modeling ability for object detection. The performance gap of mainstream SSL methods
in image classification and object detection is as suggested in Figure \ref{fig:1}. It can be observed that the linear classification accuracy of the relevant SSL methods on ImageNet dataset is increasing from 67.5\% to 75\%. While compared with MOCO~\cite{he2020momentum}, which is proposed in 2020, the detection performance (mAP) of recent approaches is  reduced when finetuning on MSCOCO dataset.

\begin{figure*}[t]
  \includegraphics[width=\textwidth]{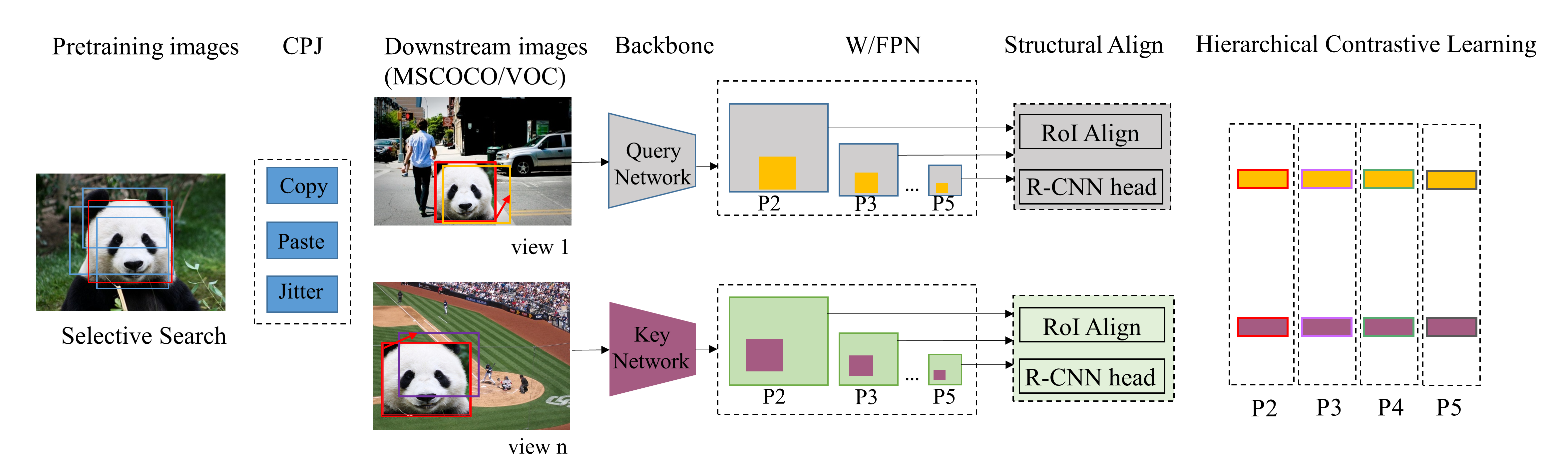}
  \caption{The overview of CoDo. For a pretraining image, we utilize selective search to generate  a series of proposals. Only one proposal is randomly selected, then pasted onto various downstream background images. } 
  
  \label{fig:2}
\end{figure*}

Existing works that contribute to bridge the gap usually focus on pretext task and architectural alignment. Firstly, instance discrimination, the typical SSL pretext task, usually assumes that different data-augmentations (views) of the same image should be similar but discriminable from other images. Generally speaking, researchers tend to believe that image-level pretext task is not suitable for object-level task. Namely, instance discrimination is suit for image classification datasets, such as ImageNet~\cite{deng2009imagenet}, which is usually single-centric-object. However, not for object detection datasets, which mainly consist of non-iconic images, such as MSCOCO~\cite{lin2014microsoft}, there are  multiple instances on an image. And secondly, the spatial modeling required for object detection can be optimized during self-supervised pretraining by aligning model structures, such as introducing feature pyramid network~\cite{lin2017feature}, RoIAlign~\cite{he2017mask} and so on. Although SSL makes it possible to involve downstream datasets during pretraining,  we identify that prior works still overlook the function of downstream datasets. Specifically, obtaining the location ability for same foreground objects of pretraining datasets, when the various background images are from object detection datasets.

Motivated by this, we present a new object-level contrastive learning method that fuses pretraining and downstream datasets, called \textbf{Co}ntrastive learning with \textbf{Do}wnstream background invariance (CoDo). We firstly generate object proposals for pretraining images by selective search~\cite{uijlings2013selective}, and paste them at arbitrary aspect ratios and scales onto various downstream background images. Then, by introducing bounding box jitter, proposals with background information will be regarded as views for object-level contrastive learning.

Involving bounding box in pretraining, we are allowed to refer to the object detectors for structural alignment. Apart form pre-training the  backbone,  our approach realizes a better initialization for all components in detectors. Experimentally, we study mainstream detection backbone network, ResNet50-FPN on  MSCOCO. Our approach shows impressive improvements over the baseline.


\section{Related work}

Self-supervised learning refers to learn visual features from unlabeled data without human annotations~\cite{jing2020self}. A mainstream solution is to utilize various pretext tasks to generate pseudo labels in order to obtain the generalizable representations, such as Rotation~\cite{gidaris2018unsupervised}, Colorization~\cite{zhang2016colorful}, Inpainting~\cite{pathak2016context}, Jigsaw Puzzle~\cite{doersch2015unsupervised} and so on. Recently, contrastive learning  becomes the most popular image-level pretext task for self-supervised learning~\cite{chuang2020debiased}. The optimization goal is to maximize the representation of image instances and their corresponding views, while minimizing the remaining other image instances~\cite{wu2018unsupervised}. The further contrastive learning approaches  focus on a better construction of negative sample and a simplification of network structure. For instance, Momentum Contrast (MoCo)~\cite{he2020momentum} builds a moving-averaged encoder and maintain a negative sample queue to minimize a contrastive loss. SimCLR~\cite{chen2020simple} verifies the effectiveness of data augmentation strategy and increases the batch size of training samples for contrastive learning. Bootstrap Your Own Latent (BYOL)~\cite{grill2020bootstrap} minimizes a similarity loss between online and target networks without using negative pairs. Barlow Twins~\cite{zbontar2021barlow} introduces a cross-correlation matrix to avoid trivial constant solutions, the simplified model dose not need predictor network, momentum encoder and  stop-gradients any more.

The further goal of SSL is to learn general representations which can be transferred for downstream tasks. Some modified approaches have been proposed to bridge the gap between pretraining and downstream tasks. Detco~\cite{xie2021detco} and Insloc~\cite{yang2021instance} separately design a detection-friendly pretext task. Detco contrasts the global image and local image patches to improve object detection. Insloc constructs a new view by randomly pasting foreground objects in different background images for contrastive learning. OLR~\cite{xie2021unsupervised} and Soco~\cite{wei2021aligning} propose object-level unsupervised representation learning framework for object detection respectively. Self-EMD~\cite{liu2020self} performs the Earth Mover’s Distance as a metric to measure the spatial similarity between two image representation in order to learn spatial visual representations for object detection. DiLo ~\cite{zhao2020distilling} selects saliency estimation to localize the foreground object in a data-driven approach. Self-EMD~\cite{liu2020self} directly pre-trains on MSCOCO and achieves a higher detection performance, instead of commonly used ImageNet.

\section{Method}
We propose a new object-level contrastive learning framework as shown in Figure \ref{fig:2}, which introduces the unlabeled downstream datasets information during pre-training. The proposals of the pretraining images are pasted to the downstream images, the bonding box of proposals is deformed to generate new views with downstream information.  It enables localization of the foreground object during pretraining, and can be regarded as a simulation of object detection during pretraining. Considering that no supervised information of downstream dataset is introduced into the pre-training, so there is no risk of data leakage. Meanwhile, we introduce architectural
alignment to pretrain the essential properties of object detector, such as FPN, RoiAlign and R-CNN head. Our proposed method is detailed below.

\subsection{Copy, paste and jitter (CPJ)}
We design a data augmentation method for proposals to realize location modeling, terms as CPJ. In order to realize object-level contrastive learning, we select selective search to generate proposal in unsupervised way. Considering ImageNet is usually regarded as a single-centric-object dataset, most proposals are similar. So we only randomly select one proposal for each image, those proposals with too large or too small aspect ratios ($>=3$ or $<=1/3$) are ignored. In addition to ImageNet, we add two  downstream object detection datasets (MSCOCO and Pasco VOC~\cite{Everingham15}) as alternative background images. For the same pretraining image, depending on the number of 
views involved in the contrastive learning, we randomly select the corresponding number of background images to generate the pasted images.

Background invariance is important for object detection, namely, a robust detector can recognize the foreground objects on various backgrounds. We paste the proposals at arbitrary aspect ratios and scales onto various downstream background images. In this process, translation and scale invariance are also considered. The pasted position is treated as the bounding box. Then the bounding box is jittered to contain background images. The jittered boxes  are filtered by a proper IoU threshold,  we set it to be greater than 0.6.  The whole image $I$ and transformed bounding box $bb$   are regarded as inputs of Query network and Key network to conduct 
contrastive learning.

\begin{equation}
I_{q}, bb_{q}=CPJ(I_{p},I_{dq}).
\end{equation}
\begin{equation}
I_{k_i}, bb_{k_i}=CPJ(I_{p}, I_{d{k_i}}).
\end{equation}
Where $I_{p}$ is the proposal of pretraining images, $I_{dq}$ is the downstream background image for Query network, and $I_{d{k_i}}$ is the i-th downstream background image for Key network.

To be consistent with existing SSL methods, the default version of our proposed method selects two views for Query network and Key network separately. Actually, multi views help to increase diversity during contrastive learning. We also design a 4 views version, which view 2 to view 4 are inputs of Key network. 

\subsection{Hierarchical contrastive learning}

In our approach, the pipeline of MOCO-V2~\cite{chen2020improved} is adopted as baseline for learning contrastive representations. We will describe how to achieve structural alignment for MOCO V2 in this section. We select the Resnet50 with FPN as Query network $f^q$  and Key network $f^k$. FPN is a common component in the object detector, which can fuse different feature maps to reach a better detection performance. 
To further align with Mask R-CNN, RoIAlign is introduced to extract feature of bounding box from the output of FPN. For Query network, the object-level feature representation $v_q$ is extracted from an image $I_q$ and the corresponding bounding box $bb_q$ as follows. The computation for Key network is similar.

\begin{equation}
v_{q}=\text { RoIAlign }\left(f^{q}\left(I_{q}\right), bb_{q}\right).
\end{equation}

\begin{equation}
v_{k_i}=\text { RoIAlign }\left(f^{k}\left(I_{k_i}\right), bb_{ki}\right).
\end{equation}

And R-CNN head $f^{R-H}$ is built to obtain embeddings for contrastive learning.
The latent embeddings $e_{q}$ and $e_{k_i}$ for Query network  and Key network are as follows:
\begin{equation}
e_{q}=f^{R-H}(v_{q}).
\end{equation}

\begin{equation}
e_{k_i}=f^{R-H}(v_{k_i}).
\end{equation}

\begin{table*}[t]
\footnotesize
  \caption{Downstream task performance on COCO by using Mask R-CNN with R50-FPN.}
  \label{tab:1}
  \begin{tabular}{c|c|cccccc|cccccc}
    \toprule
    \multicolumn{1}{c|}{\multirow{2}{*}{Methods}} &\multicolumn{1}{c|}{\multirow{2}{*}{Epoch}} & \multicolumn{6}{|c|}{1x Schedule} & \multicolumn{6}{|c}{2x Schedule} \\
    \multicolumn{1}{c|}{}  &\multicolumn{1}{c|}{} & $AP^{bb}$& $AP^{bb}_{50}$& $AP^{bb}_{75}$ &$AP^{mk}$ &$AP^{mk}_{50}$ &$AP^{mk}_{75}$ &$AP^{bb}$ & $AP^{bb}_{50}$& $AP^{bb}_{75}$ & $AP^{mk}$ &$AP^{mk}_{50}$&$AP^{mk}_{75}$\\
    \midrule
    \texttt \ Sepurvised  & 90& 38.9 &59.6  & 42.7& 35.4 &56.5 &38.1 &41.3  &61.3  & 45.0  &37.3& 58.3& 40.3\\
    \texttt \ MOCO & 200&38.5& 58.9& 42.0& 35.1& 55.9& 37.7&40.8& 61.6& 44.7& 36.9& 58.4& 39.7\\
    \texttt \ MOCO v2 &200& 40.4& 60.2& 44.2& 36.4& 57.2& 38.9&41.7& 61.6& 45.6& 37.6& 58.7& 40.5\\
    \texttt \ SimCLR  & 200 &39.6& 59.1& 42.9& 34.6& 55.9& 37.1& 40.8& 60.6& 44.4& 36.9& 57.8& 39.8\\
    \texttt \ InfoMin & 200	&40.6& 60.6& 44.6& 36.7& 57.7& 39.4& 42.5& 62.7& 46.8& 38.4& 59.7& 41.4\\
    \texttt \ InfoMin & 800	&41.2& 61.2& 44.8& 35.9& 57.9& 38.4 &42.1 &62.3& 46.2& 38.0& 59.5& 40.8\\
    \texttt \ BYOL  &300 &40.4& 61.6& 44.1 &37.2& 58.8& 39.8& 42.3& 62.6& 46.2& 38.3& 59.6& 41.1\\
    \texttt \ SWAV &400 &39.6& 60.1& 42.9& 34.7& 56.6& 36.6&42.3& 62.8& 46.3& 38.2& 60.0& 41.0\\
    \texttt \ DenseCL & 200& 40.3& 59.9& 44.3& 36.4& 57.0& 39.2& 41.2& 61.9& 45.1& 37.3& 58.9& 40.1\\
    \texttt \ InsLoc& 200& 41.4& 61.7& 45.0& 37.1& 58.5& 39.6& 43.2& 63.5& 47.5& 38.7& 60.5& 41.9\\
    
    \textbf{CoDo }   & 200	&41.2	&{61.3}	&{45.1}	&{36.9	}&58.2	&39.4	&42.6&	{62.6}&	{46.7}&{	38.2} &59.7 &41.0\\
     
     \textbf{CoDo }   & 400	&41.9	&{61.8}	&{45.8}	&{37.4	}&58.8	&40.3	&43.1&	{63.4}&	{47.1}&{38.8} &60.6 &41.4\\
     
      \textbf{CoDo$_{m}$ }   & 400	&43.1	&63.3	&47.1	&38.3&60.3	&41.2	&-&	-&	-&- &- &-\\
					
    \bottomrule
  \end{tabular}
\end{table*}

\begin{table*}\centering
\footnotesize
  \caption{The influence of background datasets on COCO by using Mask R-CNN with R50-FPN. Q means Query Network and K means Key Network.}
  \label{tab:2}
  
  \begin{tabular}{c|ccccccccccc}
    \toprule
     Background Datasets &Epoch&Schedule& $AP^{bb}$ & & $AP^{bb}_{50}$& $AP^{bb}_{75}$ &$AP^{mk}$  &$AP^{mk}_{50}$ &$AP^{mk}_{75}$ 	\\
    \midrule
    \texttt \ Q:ImageNet K: COCO &200&1x& 40.7&   --    &60.7&	44.5&	36.5	&57.6&	39.1\\
    \texttt \ Q/K: ImageNet+COCO  &200&1x& 40.8   & +0.1  & 60.7&	44.7&	36.6&	57.6&	39.3\\
    \texttt \ Q/K: ImageNet+COCO+VOC &200&1x& 41.2&	+0.5&61.3&	45.1&	36.9&	58.2&	39.4\\
    \texttt \ Q: ImageNet K:ImageNet+COCO+VOC &400&1x& 41.4&--	&61.7&	45.1&	37.3&	58.6&	40.1\\
    \texttt \ Q/K: ImageNet+COCO+VOC &400&1x& 41.9&	+0.5&61.8&	45.8&	37.4&	58.8&	40.3\\

    \bottomrule
  \end{tabular}
\end{table*}

SSL usually selects contrastive loss, i.e. InfoNCE to compute the similatity between views. For Query view and i-th Key view, the loss function is as follows. Notably, The calculation is performed in a hierarchical manner. Because $e_{q}$ and $e_{k_i}$  can be divided corresponding to the output of FPN $(P2,P3,P4,P5)$. Contrastive learning can be carried out at a finer level.

\begin{equation}
L_{q-k_{i}}=-\log \frac{\exp (e_{q} \cdot v_{e_{i}} / \tau)}{\sum_{i=0}^{N} \exp \left(e_{q} \cdot e_{k_{i}} / \tau\right)}
\end{equation}
Where  $\tau$ and $N$ are the temperature and the number of negative samples, respectively.

For multi-view version of our proposed method, we can calculate InfoNCE between view q and all view k. The total loss function is as follows.

\begin{equation}
L_{q-k}=\sum_{j=0}^{N}  (L_{q-k_{i}})_{j}
\end{equation}

Where $N$ is the  number of view k.

\section{Experiment}
\subsection{Dataset}
The widely used ImageNet-1K with 1.28 million images is adopt as dataset for self-supervised pretraining.  MSCOCO is used for finetuning to evaluate the generalization performance on downstream task. Significantly, in the procession of data processing, the training sets of ImageNet-1K, PASCAL 
VOC0712 and MSCOCO are involved as the background images.
\subsection{Setting for pretraining and finetuning}
During pretraining, we mainly follow the hyper-parameters setting of Moco-v2, the total batch size is set to 1024 over  8 Nvidia A100 GPUs, and the initial learning rate is set to 0.06. The optimization takes 200 and 400 epochs for the evaluation of downstream tasks, respectively. During pretraining, we employ the data augmentation pipeline of Moco-v2 for pretraing proposals and the background images.

For finetuing, we validate the performance of pre-training representation on downstream tasks based on Detectron2~\cite{wu2019detectron2}. On COCO, we adopt the Mask R-CNN with the R50-FPN. The performances of object detection and instance segmentation under 1× and 2× schedules are reported. The batch size is set to 128 over 8 GPUs. The fine tuning iteration step is set to 90000 and 180000 on the 1x schedule and 2x schedule, respectively. The initial learning rate is 0.02. Finally, on the 1x schedule and 2x schedule, the results of$AP^{bb}$,$AP^{bb}_{50}$ and $AP^{bb}_{75}$ for object detection, $AP^{mk}$, $AP^{mk}_{50}$, and $AP^{mk}_{75}$ for instance segmentation are compared with the state-of-the-art methods. R50-FPN (ResNet-50 with FPN) is the common backbone network for Mask R-CNN and Faster R-CNN to evaluate transfer performance.

\subsection{Results}
We report the experimental results on object detection and instance segmentation  with state-of-the-art approaches. Some of these sota methods are designed to focus on classification, such as SimCLR~\cite{chen2020simple}, MoCo~\cite{he2020momentum}, MoCo v2~\cite{chen2020improved}, BYOL~\cite{grill2020bootstrap} , InfoMin~\cite{tian2020makes} and SwAV~\cite{caron2020unsupervised}. While others are specifically suited to object detection, such as DenseCL~\cite{wang2021dense} and InsLoc~\cite{yang2021instance}.

\textbf{Mask R-CNN on MSCOCO.} Table~\ref{tab:1} shows the results for Mask R-CNN with R50-FPN. The finetuning follows with the COCO 1× and 2× schedules. We compare the proposed method under 200 and 400 epochs of pretraining. The results show that our method exceeds the above two kinds of methods. On 1× schedule,  Our method outperforms the baseline MoCo-v2 by +0.8  AP for the R50-FPN. On 2× schedule, our method exceeds MoCo-v2 by +0.8. The multi-view version of CoDo$_{m}$ further boosts the performance and reaches a 43.1 AP.

\textbf{The influence of Background datasets.} We analyze the background datasets selection strategy for Query Network and Key Network.
Table~\ref{tab:2} shows the influence of background datasets on object detection. Firstly, considering the volume gap between ImageNet and object detection datasets, if we only select the object detection datasets as background may cause homogenization. A proper solution is to gather ImageNet and object detection datasets together as the source of the background images. And the diversity of background dataset sources helps to improve performance. Under 200 epochs, this setting contributes a +0.5AP. Secondly, the background datasets selection for Query and Key Network should be same to reduce Disequilibrium. Under 200 epochs, this role also contributes a +0.5AP.

\section{Conclusion}
In this paper, we noticed that most existing self-supervised learning methods ignore the function of downstream datasets, especially the location ability of foreground objects in downstream scenes. So we propose a new contrastive learning method, CoDo, to achieve the downstream background invariance of the foreground objects. It is implemented by pasting foreground proposals onto various downstream images for contrastive learning. CoDo achieves a strong results on transfer performance for object detection on  MSCOCO. The experimental results demonstrate that transfer performance for object detection can be strengthened by considering downstream background invariance.









{\small
\bibliographystyle{ieee_fullname}
\bibliography{egbib}
}

\end{document}